\documentclass[12pt,a4paper]{article}

\usepackage{fullpage}
\usepackage{authblk}
\usepackage[english]{babel}
\usepackage[utf8x]{inputenc}
\usepackage{amsmath}
\usepackage{amssymb}
\usepackage{graphicx}
\graphicspath{ {images/} }
\usepackage{longtable}
\usepackage{multirow}
\usepackage{hyperref}
\usepackage{subcaption}

\makeatletter
\newcommand*\bigcdot{\mathpalette\bigcdot@{.5}}
\newcommand*\bigcdot@[2]{\mathbin{\vcenter{\hbox{\scalebox{#2}{$\math#1\bullet$}}}}}
\makeatother

\usepackage{array}
\newenvironment{conditions}
{\par\vspace{\abovedisplayskip}\noindent\begin{tabular}{>{$}l<{$} @{${}={}$} l}}
	{\end{tabular}\par\vspace{\belowdisplayskip}}

\providecommand{\keywords}[1]{\textbf{\textit{Index terms---}} #1}

\title{VFPred: A Fusion of Signal Processing and Machine Learning techniques in Detecting Ventricular Fibrillation from ECG Signals}

\author[1]{Nabil Ibtehaz}
\author[1]{M. Saifur Rahman}
\author[1,*]{M. Sohel Rahman}

\affil[ ]{ 1017052037@grad.cse.buet.ac.bd \\ \{mrahman,msrahman\}@cse.buet.ac.bd}

\affil[1]{Department of CSE, BUET,\protect\\ ECE Building, West Palasi, Dhaka-1205, Bangladesh}
\affil[*]{Corresponding author}

\begin{document}
\maketitle

\begin{abstract}

Ventricular Fibrillation (VF), one of the most dangerous arrhythmias, is responsible for sudden cardiac arrests. Thus, various algorithms have been developed to predict VF from Electrocardiogram (ECG), which is a binary classification problem. In the literature, we find a number of algorithms based on signal processing, where, after some robust mathematical operations the decision is given based on a predefined threshold over a single value. On the other hand, some machine learning based algorithms are also reported in the literature; however, these algorithms merely combine some parameters and make a prediction using those as features. Both the approaches have their perks and pitfalls; thus our motivation was to coalesce them to get the best out of the both worlds. Hence we have developed, VFPred that, in addition to employing a signal processing pipeline, namely, Empirical Mode Decomposition and Discrete Fourier Transform for useful feature extraction, uses a Support Vector Machine for efficient classification. VFPred turns out to be a robust algorithm as it is able to successfully segregate the two classes with equal confidence (Sensitivity = 99.99\%, Specificity = 98.40\%) even from a short signal of 5 seconds long, whereas existing works though requires longer signals, flourishes in one but fails in the other.
\end{abstract}


\clearpage

\section{Introduction}
\label{Introduction}

Ventricular Fibrillation (VF) is a type of cardiac arrhythmia which occurs when the heart quivers instead of pumping due to some disturbance in electrical activity in the ventricles \cite{priori2002clinical}. This arrhythmia may result in a cardiac arrest leaving the patient unconscious without any pulse. Ventricular Fibrillation is found initially in about 10\% of people in cardiac arrest \cite{baldzizhar2016ventricular} and sudden cardiac arrest is responsible for approximately 6 million deaths in Europe and in the United States \cite{li2012algorithm}. Therefore, fast and accurate detection of Ventricular Fibrillation can save a lot of lives. Electrocardiograph (ECG) signal captures the electrical activities of the human heart. Trained, experienced doctors can analyze these ECG signals and understand the heart condition. However, for the detection and prevention of sudden cardiac arrests caused by arrhythmias like Ventricular Fibrillation, a continuous monitoring of the ECG signal of a patient is essential; this would require a specialized doctor to examine the ECG signal relentlessly. Unfortunately, it is neither practical nor possible for a doctor to continuously monitor ECG signals of a number of
8
 patients. This is the primary motivation for developing algorithms to analyze patterns from ECG signals of patients to detect arrhythmias.

Ventricular Fibrillation, being one of the most severe life-threatening arrhythmias, has been studied diligently for over four decades. A number of methods have been proposed for the detection of Ventricular Fibrillation over this time period. Very early works include simple signal processing \cite{thakor1990ventricular,chen1987ventricular,kuo1978computer,barro1989algorithmic,zhang1999detecting}. Subsequently, more advanced signal processing based methods were introduced \cite{arafat2009detection,anas2011exploiting,amann2005new,amann2007detecting}. Unfortunately, these methods fail to maintain accuracy when tested on a large dataset \cite{amann2005reliability} (further explained in section \ref{Comparison}). In recent past, with the emergence of machine learning techniques, better performing algorithms were introduced that extracted features from previously established ECG parameters and employed a machine learning classifier \cite{alonso2012feature,alonso2014detection,li2014ventricular,verma2016detection,song2005support,asl2008support,clayton1994recognition}. Though the performance improved, still it is far from perfection and several algorithms have certain limitations. A more comprehensive literature review is presented in section \ref{Comparison}, where we perform a comparison with other works.


The signal processing based algorithms for VF prediction actually originate from intuition, observation, and experience of the doctors; then the monumental mathematical tools of signal processing are used to exploit those patterns, and finally based on a threshold or two the decision is made. On the contrary, machine learning based algorithms treat signal characteristics as a black box, they take a number of ECG parameters (collected from those signal processing based algorithms) and train a classification model using those as features. These are more robust as learning from the pattern of a huge variety of data can potentially outperform simple and constrained rule based approaches. Hence our motivation was to blend the two: we started with the intuition of the doctors, formulated those observations using signal processing methods and finally followed a machine learning protocol to acknowledge the diversity in real medical data. Our algorithm is based on the fact that QRS complexes are absent in VF class ECG signals\cite{jones2009ecg}. We exploited this property and extracted this pattern using EMD (Empirical Mode Decomposition) with DFT (Discrete Fourier Transform) based feature engineering. We select the best set of features using Random Forests, and after making several attempts with Logistics Regression, Random Forest, Neural Networks, we finally train a SVM model and evaluate our model on benchmark datasets.

Notably, there exist a plethora of prior works exploiting EMD, DFT, and SVM, albeit mostly as separate methods. Our current work, VFPred, stands out in this context as we have been able to make a proper fusion of the methods from different domains. This is evident from the remarkable performance of VFPred as reported in a later section. More specifically, the prior SVM based works treated feature engineering as a black box and exercised a mix-and-match strategy on some arbitrary parameters. On the contrary, we have used intuition and rationale to investigate the effectiveness and efficacy of various features at length, followed by justification through extensive and thorough experimentation to attain the true potential of our chosen classifier algorithm. Also to the best of our knowledge, no prior work made an ensemble of EMD and DFT to interpret the patterns of the ECG signals.


This paper makes the following key contributions:

\begin{itemize}
  \item It proposes an elegant feature engineering scheme incorporating Empirical Mode Decomposition (EMD) with Discrete Fourier Transform (DFT), instead of using the common ECG parameters (section \ref{emd}, section \ref{dtft}).
  \item It constructs a robust SVM classifier, VFPred, that can classify both the classes with near equal high performance (section \ref{svm1}, section \ref{svm2}).
  \item It establishes the importance of feature selection and through appropriate feature ranking and selection technique increases efficiency by reducing the number of features (section \ref{Feature Ranking by Random Forest Classifier}).
  \item It adopts machine learning techniques for overcoming imbalance in the dataset, unlike downsampling by selecting convenient samples (section \ref{Overcoming the Imbalance in the Dataset}).
\end{itemize}

\section{Dataset}

For developing and evaluating VFPred algorithm, we used the following two benchmark datasets of Ventricular Fibrillation detection as has been commonly used in most other works.

\begin{enumerate}
  \item The MIT-BIH Malignant Ventricular Arrhythmia Database (VFDB) \cite{greenwald1986development}: This database includes 22, 30 minutes long ECG recordings of subjects who experienced episodes of sustained Ventricular Tachycardia, Ventricular Flutter, and Ventricular Fibrillation.
  
  The ECG signals of this database were sampled with a sampling frequency of 250 Hz.
  \item Creighton University Ventricular Tachyarrhythmia Database (CUDB) \cite{nolle1986crei} : This database includes 35, 8 minutes long ECG recordings of human subjects who experienced episodes of sustained Ventricular Tachycardia, Ventricular Flutter, and Ventricular Fibrillation. 
  
  The ECG signals of this database were filtered by an active second order Bessel low-pass filter with a cutoff frequency of 70 Hz. Then they were sampled at 250 Hz and quantized with 12-bit resolution over 10 V range.
  
\end{enumerate}

These datasets are hosted on PhysioNet \cite{goldberger2000physiobank}, and are publicly available.

Following the practice in the literature, we experimented on ECG episodes of length, $T_e = 2$ sec, 5 sec and 8 sec. In order to extract the ECG episodes, first a window of length $T_e$ sec is taken, and the window is shifted by 1 sec to collect the consequent episodes, till the end of the signal. Finally, the episodes were annotated as `VF' or `Not VF' using the expert annotations provided in the dataset. We considered the entire dataset except for the few segments annotated as noise. We only used channel 1 data from VFDB to avoid redundancy. Notably, the dataset is highly imbalanced: we roughly have only 9\% of the data as VF (Please refer to Table 1 from the supplementary material).

\section{Proposed Algorithm}
\label{Methods}
The proposed VFPred algorithm takes an ECG signal of $T_e$ seconds long, and through the following pipeline predicts whether Ventricular Fibrillation (VF) is present or not.

\subsection{Signal Preprocessing and Filtering}

ECG signals are usually corrupted by various kinds of noises and interferences.  To overcome this, the input ECG signal is preprocessed following the filtering process originally proposed in \cite{amann2005reliability} and slightly modified in \cite{anas2011exploiting}. The modified filtering procedure is as follows:

\begin{enumerate}
    \item The mean value is subtracted from all the samples, thus making the mean zero.
    \item Next, a moving average filter of order 5 is applied. For an ECG signal, this should remove most of the interspersions and muscle noise.
    \item Then, the signal is filtered using a high pass filter of cut-off frequency, $f_c = 1$ Hz. This imposes drift suppression on the signal.
    \item Finally a low pass butterworth filter of order 12  and cut-off frequency, $f_c = 20$ Hz is applied to attenuate the unnecessary high frequency information.
\end{enumerate}

\begin{figure}[h]
    \includegraphics[scale=0.5]{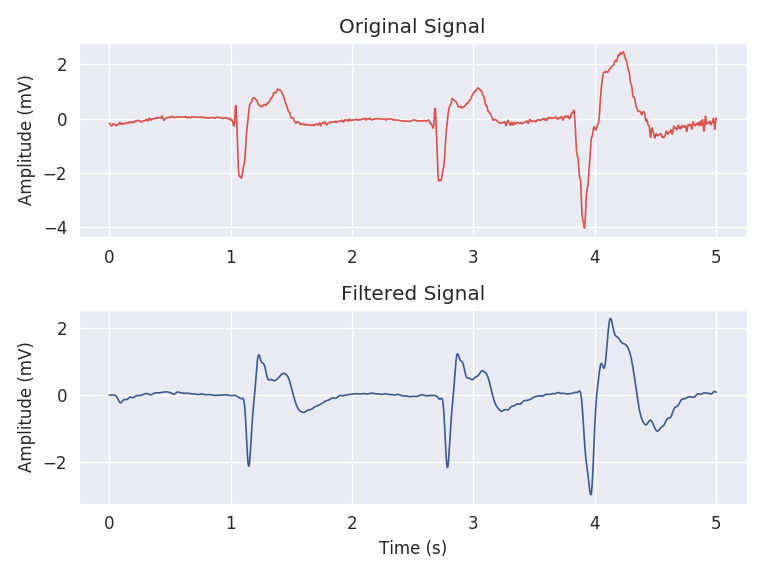}
    \centering
    \caption{Signal Preprocessing and Filtering Step. This process removes most noises and artifacts from the ECG signal and smoothens the signal. Moreover this step also makes the signal zero mean.}
    \label{fig:signalPreprocessing}
\end{figure}

Figure \ref{fig:signalPreprocessing} illustrates the outcome of preprocessing and filtering on the signal. As can be observed this step removes a great deal of noises from the ECG signal.

\subsection{Analyzing the Oscillatory Characteristics}
\subsubsection{Empirical Mode Decomposition (EMD)}
\label{emd}

Empirical Mode Decomposition (EMD), proposed by Huang et al. \cite{huang1998empirical}, is a data-driven, adaptive signal processing method which is suitable for analyzing non-stationary and nonlinear signals, like the ECG signal. This algorithm decomposes a signal into a sum of Intrinsic Mode Functions (IMF). An IMF represents a simple oscillatory function with the following characteristics:

\begin{enumerate}
    \item The number of extrema and the number of zero crossings must either be equal or differ (with each other) at most by one.
    \item At any point, the mean value of the envelopes of the local maxima and minima is zero.
\end{enumerate}

\begin{figure}[h]
    \centering
    \includegraphics[scale=0.35]{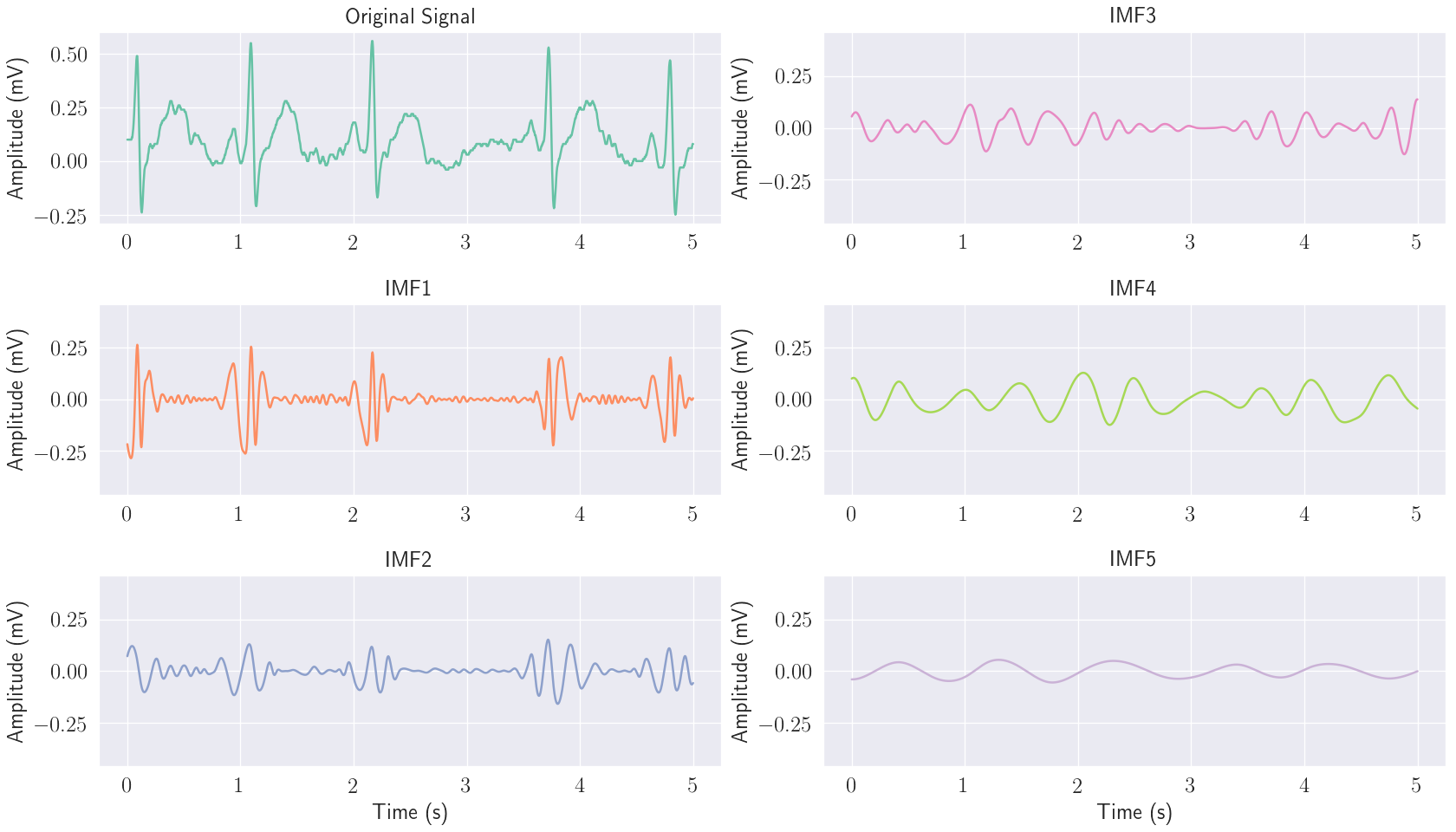}
    \caption{Empirical Mode Decomposition of an ECG Signal. Here we can observe that the top IMF components are more oscillatory in nature and gradually their oscillation diminishes. $IMF_1$ and $IMF_2$ captures most of the oscillations from the original signal and it progressively fades in  $IMF_3$, $IMF_4$ and $IMF_5$.}
    \label{fig:emd}
\end{figure}

Please refer to the supplementary materials for the complete description of the EMD algorithm. By applying the EMD algorithm we decompose a signal x(t) into a set of IMF functions in decreasing order of oscillatory behavior, and possibly a Residue, as shown in Figure \ref{fig:emd}.
\begin{equation}
x(t)=IMF_1(t)+IMF_2(t)+...+IMF_n(t)+R(t)
\end{equation}

\subsubsection{Observing IMF components to Distinguish VF from Not VF}

IMF components describe the oscillatory characteristics of a signal. From the studies of ECG signal, it has been found that except for `VF' signals all other classes of ECG signals contain a QRS complex \cite{jones2009ecg}. Thus, we have a property that apparently separates the `VF' from the `Not VF'. An attempt can be made to find a proper formulation to distinguish `VF' signals exploiting this characteristic.

The presence of a QRS complex distorts the upper and lower envelopes as they introduce additional local maxima and local minima points. On the contrary due to the absence of any QRS complex, `VF' class ECG signals have envelopes that are symmetric in nature. The small interval QRS complexes in the `Not VF' class ECG signals result in higher frequency oscillations. Thus their 1st IMF component captures those oscillations missing the actual ECG signal itself. On the other hand, for the lack of QRS complex, the `VF' class ECG signals are not supposed to have major high frequency oscillations compared to `Not VF' class ECG signals. Thus the 1st IMF component of `VF' class is likely to follow the original ECG signal.

Thus, it should be possible to distinguish the two classes by observing how much the 1st IMF component correlates with the ECG signal itself. Moreover, the remaining signal can also be analyzed to increase the robustness, which will be referred as the Residue component \cite{anas2011exploiting}. Since the IMF and Residue components are disjoint in nature it is expected that the Residue of `Not VF' signal will match the original ECG signal more than that of `VF' class ECG signal. The cosine similarity metric can be taken as a measure of similarity between the two signals \cite{anas2011exploiting}. Thus $IMF_{similarity}$ and $R_{similarity}$ can be defined as follows:

\begin{equation}
IMF_{similarity}= \frac{Signal \cdot IMF_1}{\mid Signal \mid \text{} \mid IMF_1 \mid}
\end{equation}
\begin{equation}
R_{similarity}= \frac{Signal \cdot R}{\mid Signal \mid \text{} \mid R \mid}
\end{equation}

\begin{figure}[h]
    \centering
    \begin{subfigure}[h]{0.49\textwidth}
        \includegraphics[width=\textwidth]{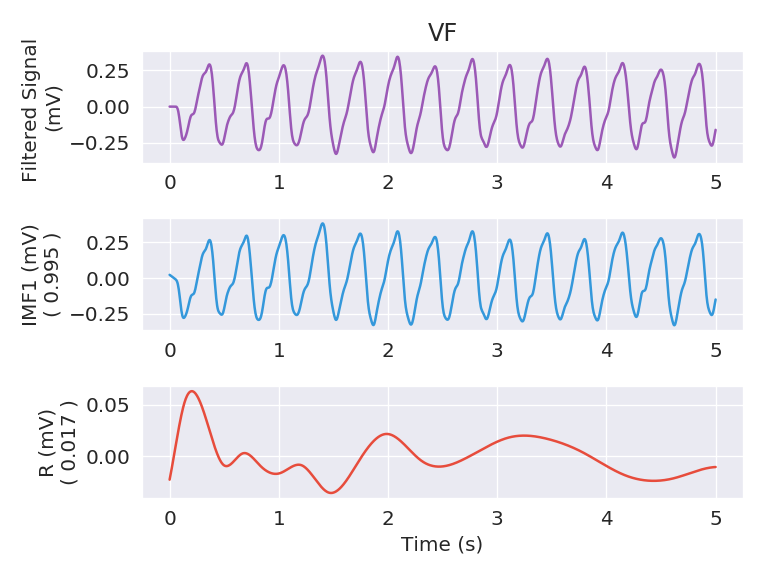}
        \caption{}
        \label{fig:VF_imfRsim}
    \end{subfigure}    
    \hfill
    \begin{subfigure}[h]{0.49\textwidth}
        \includegraphics[width=\textwidth]{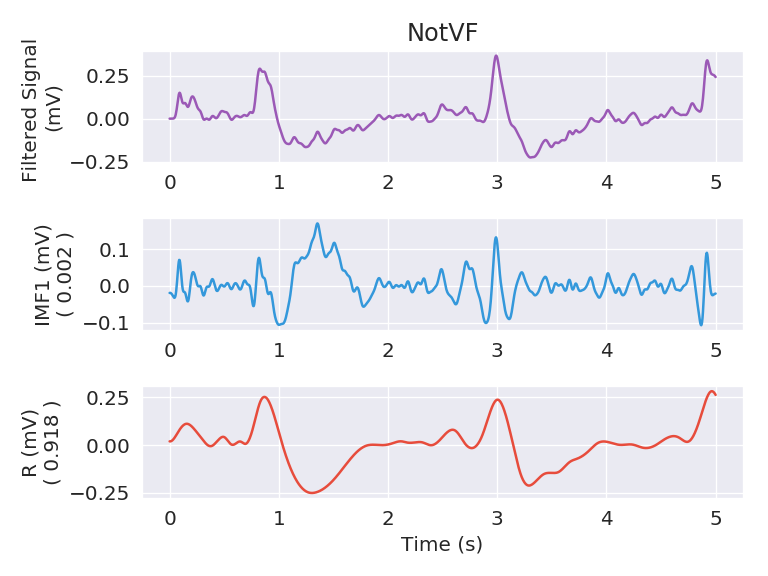}
        \caption{}
        \label{fig:NotVF_imfRsim}
    \end{subfigure}
    \caption{Relation of the 1st IMF and R components with the original signal, the cosine similarities are presented inside the parentheses. In (a) for a `VF' class signal, we observe that the 1st IMF and the original signal are very similar (cosine similarity = 0.995) while R deviates quite a bit (cosine similarity = 0.017). (b) on the other hand, shows an example from `Not VF' class. Here, we can observe distinct QRS peaks, and they prevent the 1st IMF from capturing any useful information (cosine similarity of 0.002 with the original signal). Rather they are quite oscillatory accounting for the non-uniform upper and lower envelops. This compels the residue to closely follow the original signal (cosine similarity = 0.918).}
    \label{fig:vfNotvfrimf}
\end{figure}

These facts are illustrated in Figure \ref{fig:vfNotvfrimf},

\subsubsection{Extracting Oscillatory Characteristics from ECG Signals}

Even after preprocessing and filtering, some high frequency noises may still prevail in the signal. In such cases, the 1st IMF component would actually represent those noises. To determine whether the 1st IMF component contains useful information or noise we follow the scheme proposed in \cite{anas2011exploiting}. In order to extract the oscillatory features from the ECG signals we perform the following steps:

\begin{itemize}
    \item Empirical Mode Decomposition is performed on the filtered signal and the first two IMF components along with the Residue are computed as follows:
    \begin{equation}
        x(n) = IMF_1(n) + IMF_2(n) + R(n)
    \end{equation}
    where:
    \begin{conditions}
        x(n)     &  Filtered Signal \\
        IMF_1(n) &   1st IMF component \\
        IMF_2(n) &   2nd IMF component \\
        R(n)  &   The Residue
    \end{conditions}
    
    \item The noise level ($V_n$) is calculated as a percentage of the maximum ECG signal amplitude as follows:
    \begin{equation}
        V_n= \alpha (\max(x(n)))
    \end{equation}
    Where, $\alpha$ is a constant that should be chosen wisely. We took $\alpha=0.05$.
    
    \item The samples $n_L$ of $IMF_1(n)$ that falls within $-V_n$ to $V_n$ are identified, i.e.,
    \begin{equation}
        n_L = \{ t : t \in n  , | IMF_1(t) | \leq V_n \}
    \end{equation}

    \item The noise level crossing ratio $(NLCR)$ is calculated using the following formula,
    \begin{equation}
        NLCR = \frac{\sum_{n\in n_L}IMF_1^2(n)}{\sum_{n\in n_L}x^2(n)}
    \end{equation}

    \item Finally, the appropriate IMF is selected as follows :
    
    \begin{equation}
        IMF=
    \left\{ \begin{array}{ll}
    IMF_1(n)+IMF_2(n) & \text{, if } NLCR \leq \beta \\
    IMF_1(n) &\text{, otherwise}
    \end{array} \right.
    \end{equation}
    
    Here, $\beta$ is a properly chosen constant. We considered $\beta=0.02$.
\end{itemize}

Thus, the appropriate IMF component and the Residue is taken for further analysis.

\subsubsection{Observing Oscillatory Characteristics on Dataset}
Based on our analysis above, the expected value of $IMF_{similiarity}$ for `VF' is greater than 0.5 whereas it is less than 0.5 for `Not VF'. On the contrary, the expected value of $R_{similarity}$ is just the opposite, i.e., less than 0.5 for `VF' and greater than 0.5 for `Not VF'. However, it was found that these theoretical observations are not always found to be true in practice, especially, when checked against a large amount of diverse data. In particular, a lot of overlaps are observed between the two classes (please refer to Figure 1 of the supplementary material where we illustrate such cases of overlapping using a scatter plot).

\begin{figure}[h]
    \centering
    \begin{subfigure}[h]{0.45\textwidth}
        \includegraphics[width=\textwidth]{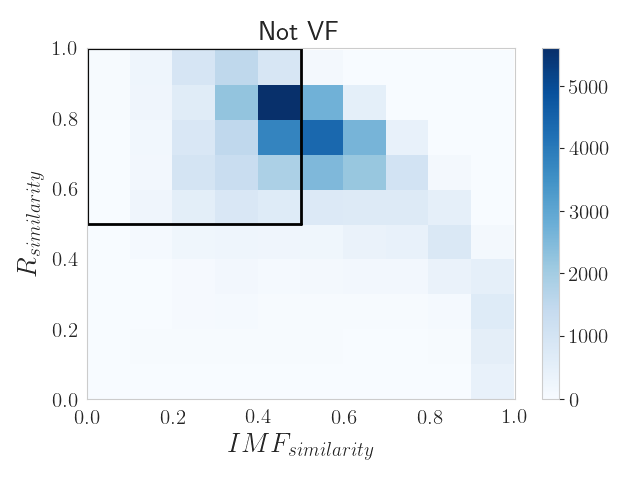}
        \caption{}
        \label{fig:notVFHist}
    \end{subfigure}    
    \hfill
    \begin{subfigure}[h]{0.45\textwidth}
        \includegraphics[width=\textwidth]{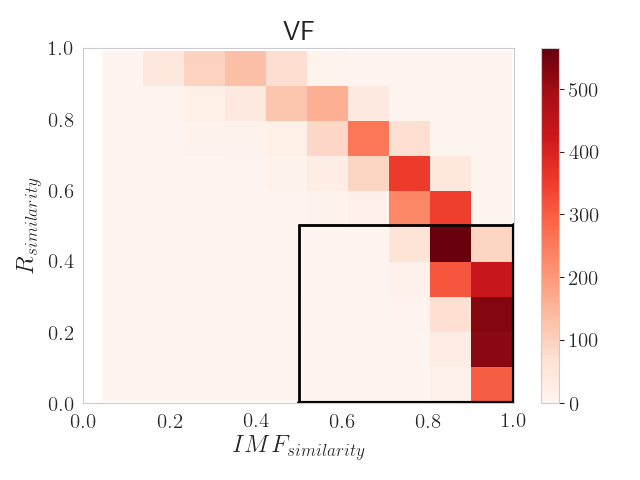}
        \caption{}
        \label{fig:VFHist}
    \end{subfigure}
    \caption{2D Histograms representing the distribution of  $IMF_{similarity}$ and $R_{similarity}$ of `VF' and `Not VF'. From our theoretical analyses, the IMF component should be similar to the original signal for a `VF' class signal and it is likely to be different for a `Not VF' class signal. On the contrary, the Residue component should diverge from the original signal for a `VF' class signal, but should closely follow the original signal for `Not VF'. Hence, bounding boxes have been drawn where the points are expected to reside. Here the two classes should have been confined within the black bounding boxes, but a lot of overlapping is clearly visible.}
    \label{fig:hist2}
\end{figure}

In Figure \ref{fig:hist2}, we have presented a 2D histogram that shows the distribution of $IMF_{similarity}$ and $R_{similarity}$ of the two classes. For both the classes, according to our theoretical analyses, the data points should have been within the bounding boxes. However, this is not the case as we can observe a great amount of overlapping.

\subsubsection{Hypothesis}

From further experimentation with data we came up with the following hypothesis:

\begin{quote}
\em Even after filtering and preprocessing, there may still remain some undesired frequency components in our data, preventing us from separating the two classes.\em
\end{quote}

In other words, there may be certain frequency components of the IMF and certain (not necessarily the same) frequency components of the Residue which may allow us to distinguish the two classes accurately. Hence, we need to prioritize these frequency components while making a decision. 

So instead of taking the straightforward cosine similarity of the Signal with IMF and R as features, as is done in \cite{anas2011exploiting}, we focus on the frequency information.

\subsection{Extracting Frequency Information from Oscillations}
\label{dtft}

\subsubsection{Discrete Fourier Transform (DFT)}
The Fourier Series proposed by Joseph Fourier in $19^{th}$ century, was originally developed to represent a periodic signal as an infinite sum of simple harmonic oscillations of different frequencies \cite{oppenheim1999discrete}. This allows us to analyze the impact of individual frequency bands over a signal. Later it was extended to aperiodic signals through Fourier Transform. For finite and discrete signals we sample the Fourier Transform coefficients as a finite sequence, that corresponds to a complete period of the original signal \cite{oppenheim1999discrete}. Thus, Discrete Fourier Transform (DFT) for a discreate signal $x$ of length $N$ is defined as:

\begin{equation}
    X[k] = \sum_{n=0}^{N-1}x[n] \exp({-\frac{2 \pi i}{N}kn}), \quad for \quad 0\leq k \leq N-1
\end{equation}

\subsubsection{Observing The Frequency Components}

Our theoretical analyses indicate that for a `VF' class ECG signal, the IMF component should be similar to the original signal, and the Residue component is likely to deviate from it. The opposite is expected for a `Not VF' class ECG signals. But in studies on real world dataset, a lot of fluctuations are observed. Thus, according to our hypothesis we resolve the Signal, IMF and Residue to frequency components using DFT, and then analyze their relations.

\begin{figure}[htbp]
	\centering
	\begin{subfigure}[h]{0.79\textwidth}
		\includegraphics[width=\textwidth]{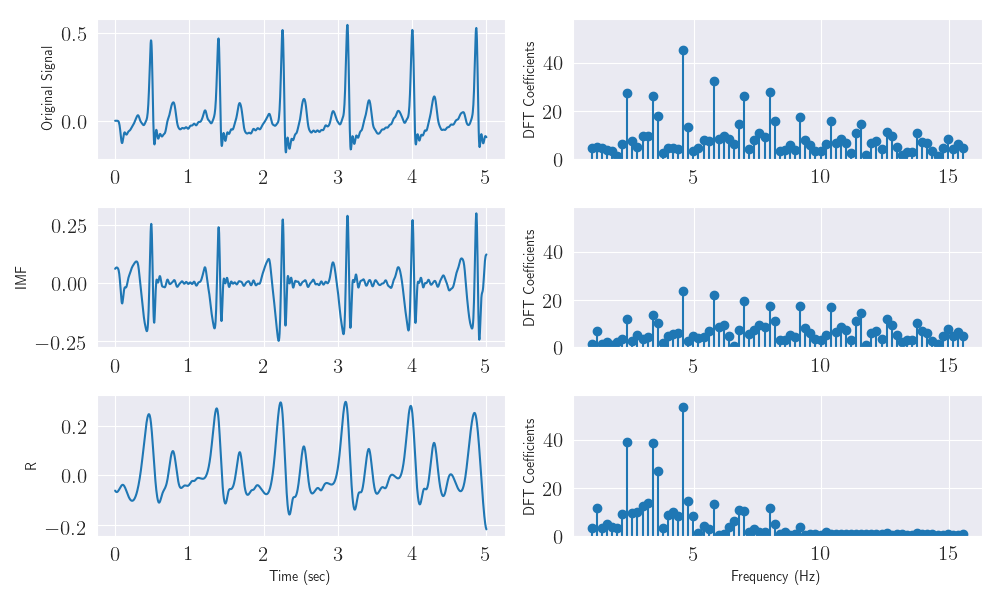}
		\caption{Not VF}
		\label{fig:dtft1}
	\end{subfigure}	
    \hfill
	\begin{subfigure}[h]{0.79\textwidth}
		\includegraphics[width=\textwidth]{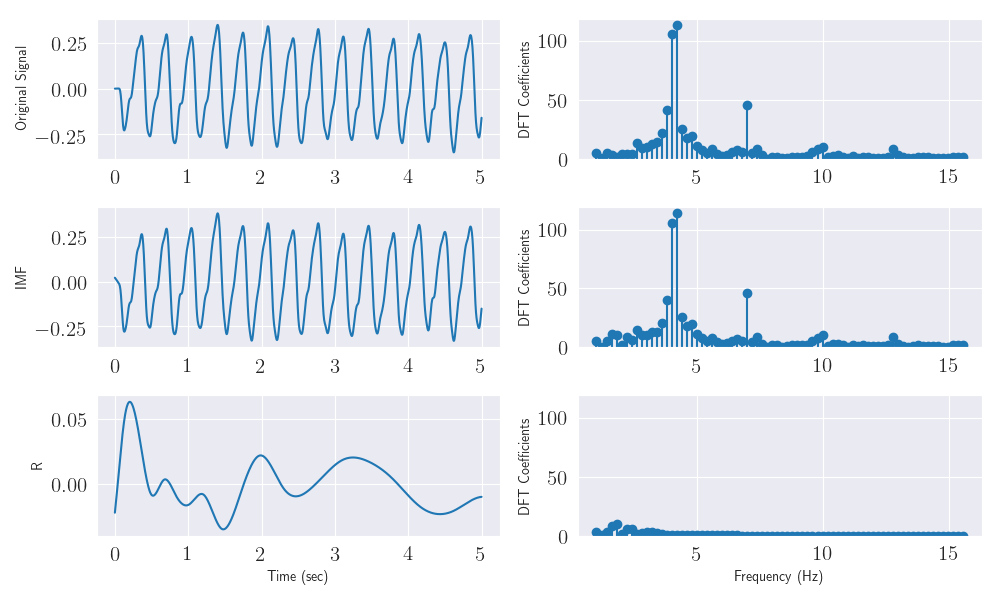}
		\caption{VF}
		\label{fig:dtft2}
	\end{subfigure}
\caption{Frequency domain observation of the two classes. In (a) for `Not VF' the DFT coefficients represent the wide band nature of the signal due to the presence of QRS complex. Here, we can observe that the IMF component accounts for the fluctuations, instead of following the signal, also the DFT coefficients appear to be representing the wide band characteristics of the signal as well. On the contrary, the R component seems to follow the pattern of the signal. On the other hand, in (b) we can observe that for `VF' class the DFT coefficients of the IMF component almost completely match with that of the signal whereas DFT coefficients of R are pretty tiny, thus they cover almost no information at all. In both the figures it can be seen that the DFT components between 1-5 Hz offer the most useful insights (here, the plots are cropped to 1-15 Hz range for visualization purposes.}
	\label{fig:dtft}
\end{figure}

In our data analysis it was found that the frequency components lying between 1 - 5 Hz were more prominent while separating the two classes. Also, if we plot the frequency components of the two classes and observe them we get a somewhat intuitive idea of classifying the signals (Figure \ref{fig:dtft}).

 In Figure \ref{fig:dtft1}, we observe the example of a `Not VF' class. Here it is evident from the DFT coefficients, that the original signal is a wide-band signal, due to the presence of QRS complex. For the case of IMF, the DFT coefficients merely captures the distributed wideband components caused by QRS complex. On the other hand, the DFT coefficients of the R signal seems to represent a quite narrow-band signal and it captures the pattern of the signal. This satisfies our expectation.

 On the other hand, in Figure \ref{fig:dtft2}, i.e., for the example of `VF' class, we can observe that the IMF frequency components, as we expected, almost completely match with that of the original signal. Additionally, a very tiny frequency components of R can be observed, which is in line with our analysis.

 In both the examples we can observe that there are some additional rather unexpected, unusual frequency components. This fact becomes more apparent as we study more data. These frequency components disturb our expected waveshape, thus affecting the validation of our assumptions. So we need to determine which frequency components should be considered. A machine learning algorithm can be used to interpret the significance of the individual frequency components and set weights accordingly to distinguish the two classes.


\subsection{Machine Learning Classifier}
\subsubsection{Feature Extraction}

Instead of taking the DFT coefficients as features, we are actually interested in how much the individual DFT coefficients of the signal matches with the corresponding DFT coefficients of IMF or R. Thus, similar to cosine similarity we multiply the two terms and normalize the product with an appropriate value.

More precisely, for both IMF and R, we multiply each of the DFT coefficients with the corresponding DFT coefficient of the original signal. Then, we normalize them by the product of the magnitude of the DFT vector of the signal with the magnitude of the DFT vector of IMF and R respectively. This gives us the similarity IMF, R and the signal in the frequency domain, which we use as features for our machine learning algorithm.

    \begin{equation}
        IMF_{similarity}[i]=\frac{Signal_{DFT}[i] \cdot IMF_{DFT}[i] }{{\mid\mid Signal_{DFT} \mid\mid \text{} \mid\mid IMF_{DFT}\mid\mid}}, \quad 1 \leq i \leq N
    \end{equation}
    
    \begin{equation}
        R_{similarity}[i]=\frac{Signal_{DFT}[i] \cdot R_{DFT}[i] }{{\mid\mid Signal_{DFT} \mid\mid \text{} \mid\mid R_{DFT}\mid\mid}}, \quad 1 \leq i \leq N
    \end{equation}

    

    
    

\subsubsection{Feature Selection using Random Forest}

By far, we have considered all the frequency components as features. However, not all features may be equally useful for our prediction and some may even hamper our prediction. This motivates us to perform a feature ranking, i.e., analyzing the importance of an individual feature.

 Random Forest \cite{ho1995random,breiman2001random} is an ensemble learning algorithm that employs a collection of decision tree classifiers. However, along with solving the classification problem, a random forest can also be used to determine the importance of the features and thereby rank them accordingly.  \cite{diaz2006gene,menze2009comparison}. 
 
 Thus using the Random Forest algorithm, we identify the most promising subset of the  features and use them for the final classification.

\subsubsection{SVM Classifier}
\label{svm1}

 Support Vector Machines (SVM), proposed by Vapnik \cite{cortes1995support,boser1992training} compute a hyperplane between the data points that separates them into two classes. Using a quadratic optimization, the hyperplane, $w^Tx + b = 0$, is constructed to maximize the distance or margin between the hyperplane and the nearest points. SVM is inherently a linear classifier but with nonlinear mappings of the input space using an appropriate kernel, SVM can be employed for nonlinear classification purposes as well. After selecting the useful features, we use an SVM classifier to classify the ECG episodes to be either `VF' or `Not VF'. We use a Gaussian Radial Basis Function (RBF) as our kernel because it reliably finds the optimal classification solutions in most practical situations \cite{keerthi2003asymptotic}. The value of a Radial Basis Function (RBF) is a function of distance from the origin ($\phi(x)=\phi(||x||)$), or some other predefined point ($\phi(x,c)=\phi(||x-c||)$). In particular, we used the Gaussian variant of RBF, which for two vectors $x$ and $x'$ is defined as follows:
\begin{equation}
    K(x,x')=\exp(-\gamma \mid\mid x-x' \mid\mid ^2)
\end{equation}
where, $\gamma$ is the inverse of the standard deviation of the gaussian distribution.

 The entire workflow of VFPred prediction system is presented in the following flow diagram:

 \begin{figure}[h]
     \includegraphics[scale=0.49]{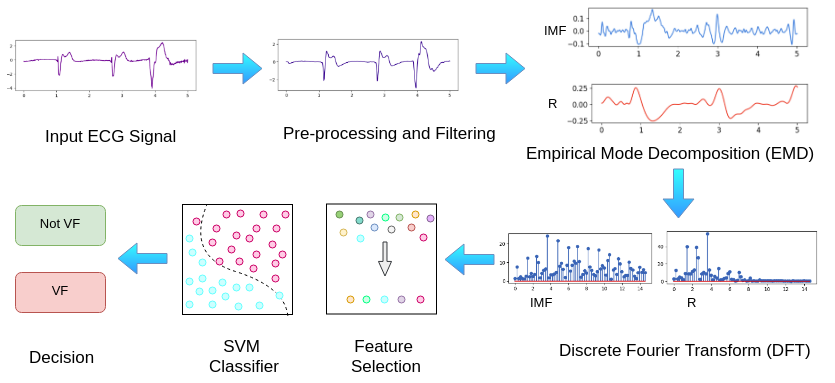}
     \centering
     \caption{Flow Diagram of VFPred Algorithm. The algorithm takes an ECG signal of length $T_e$ sec, then performs some pre-processing and filtering in order to remove noise and artifacts from the signal. Next, the IMF and R component of the signal is analyzed using Empirical Mode Decomposition. After that, the  }
     \label{fig:flow}
 \end{figure}

\section{Implementation}

We have implemented the VFPred algorithmic in Python programming language \cite{van2007python}. We used NumPy \cite{walt2011numpy} for efficient numerical computation, SciPy \cite{jones2014scipy} and PyEMD \cite{laszuk2017pyemd} for signal processing. We used SVM and Random Forest implementation from scikit-learn \cite{pedregosa2011scikit} library and used imbalanced-learn \cite{lemaitre2017imbalanced} for SMOTE . Moreover, we used matplotlib and seaborn \cite{hunter2007matplotlib} for visualization purposes. We also used the WFDB package \cite{silva2014open} to fetch data from Physionet. All the codes are available in the following github repository:

\centerline{\url{https://github.com/robin-0/VFPred}}

The experiments were performed on a Dell-Inspiron 5548 Notebook (with a 5th generation Intel core-i5 CPU @2.2 GHz having 8 GB DDR3 RAM).

\section{Evaluation Metrics}

Our problem reduces into a two class classification problem:

\begin{enumerate}
    \item Positive Class : VF
    \item Negative Class : Not VF
\end{enumerate}

We use the following commonly used evaluation metrics:
\begin{equation}
Sensitivity=\frac{TP}{TP+FN}\hspace{0.7em} Specificity=\frac{TN}{TN+FP}\hspace{0.7em}Accuracy=\frac{TP+TN}{TP+FP+TN+FN}
\end{equation}

However, since our dataset is hugely imbalanced, taking the Accuracy as the metric is not sufficient. This is because Accuracy understandably will follow the accuracy of the larger class (i.e., in our case specificity). So if an algorithm correctly identifies `Not VF' but fails to detect `VF', the Accuracy would still be high. This trend is disturbingly observed in many of the works in the literature, as most of the works prioritize the `Not VF' class \cite{anas2011exploiting,amann2005new,amann2007detecting,alonso2012feature,song2005support,asl2008support,acharya2017automated}.

Thus we need an evaluation metric that accounts for the imbalance in the dataset. Geometric Mean Accuracy (G-Mean Accuracy) is one such metric \cite{barandela2003strategies}. G-Mean Accuracy is defined as follows:

\begin{equation}
G-Mean Accuracy=\sqrt{Sensitivity \times Specificity}
\end{equation}

  This metric will become high only when both the sensitivity and specificity are high. So giving priority to the majority class as seen in some previous works will result in poor performance.

\section{Experiments}

\subsection{SVM Parameter Tuning}
\label{svm2}
Performance of any machine learning algorithm depends highly on the proper selection of parameters. For our algorithm, we have used an SVM classifier with `RBF' kernel. This SVM classifier has two parameters:

\begin{enumerate}
    \item Regularization Constant, $C$: $C$ is the parameter that defines whether the SVM classifier margin is a soft margin or a hard margin \cite{cortes1995support}.
    \item Kernel Hyperparameter, $\gamma$ : $\gamma$ is the parameter of Gaussian Radial Basis Function \cite{cortes1995support}.
\end{enumerate}

For tuning the parameters, initially, we took a small portion of the data as the training data and the rest as the validation data. Since the dataset is highly imbalanced, it was quite tricky to train our classifier. We randomly shuffled the entire dataset, and then took 3000 samples from `VF' class and 5000 samples from `Not VF' class as the training set. More samples were taken from the `Not VF' class because it contains more variations as compared to the `VF' class. The rest of the dataset was kept as the validation set (VF=2320, Not VF=46087).

\begin{figure}[tbh!]
    \centering
    \begin{subfigure}[h]{0.60\textwidth}
        \includegraphics[width=\textwidth]{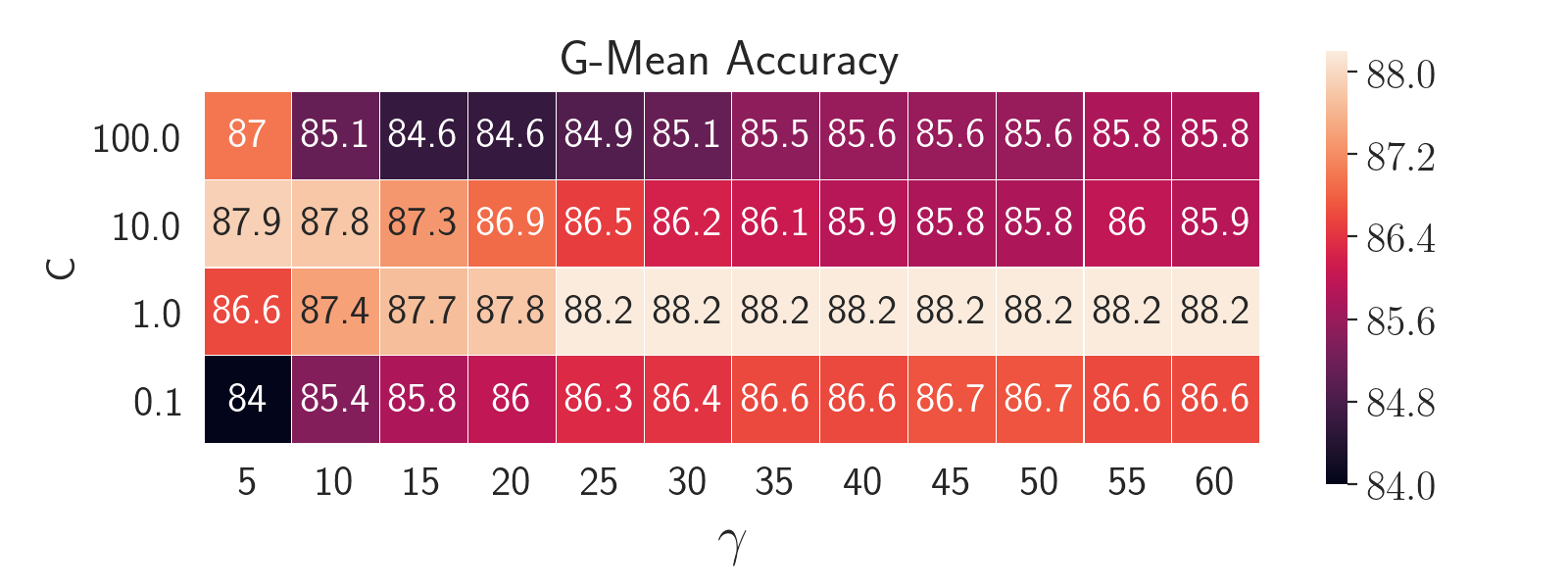}
        \caption{$T_e =2s$}
        \label{fig:svm2s}
    \end{subfigure}    
    
    \begin{subfigure}[h]{0.60\textwidth}
        \includegraphics[width=\textwidth]{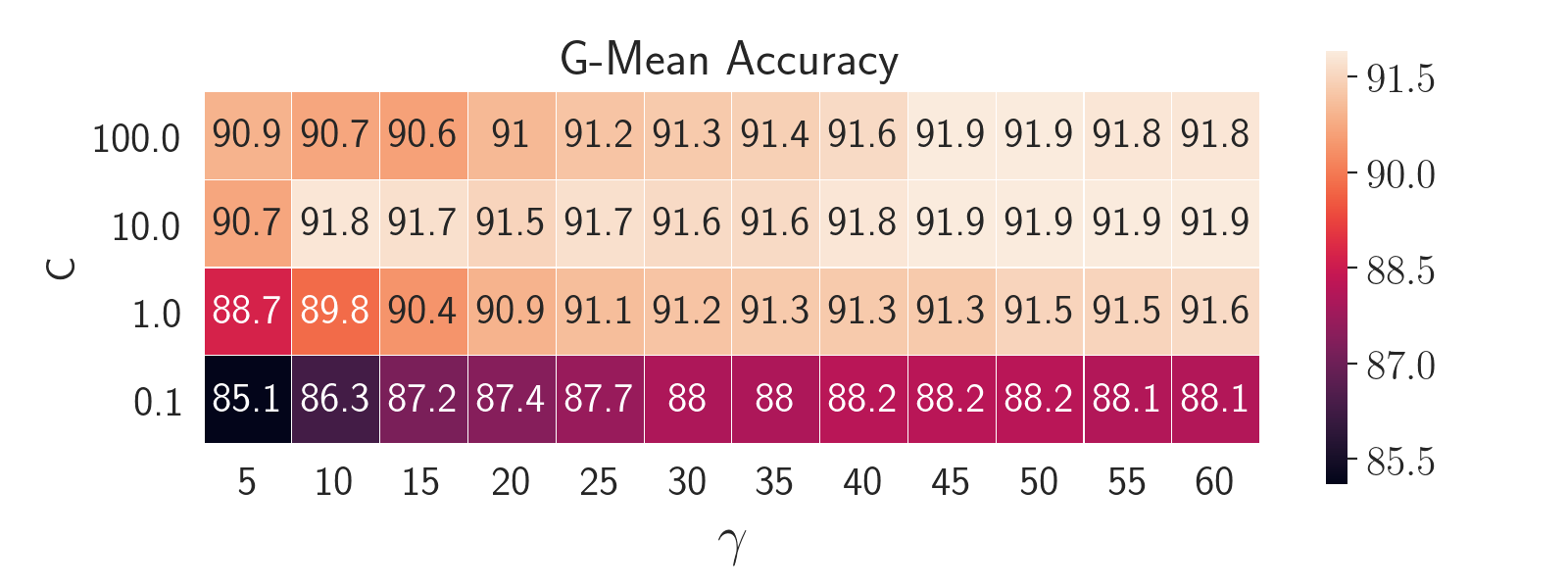}
        \caption{$T_e =5s$}
        \label{fig:svm5s}
    \end{subfigure}
    
    \begin{subfigure}[h]{0.60\textwidth}
        \includegraphics[width=\textwidth]{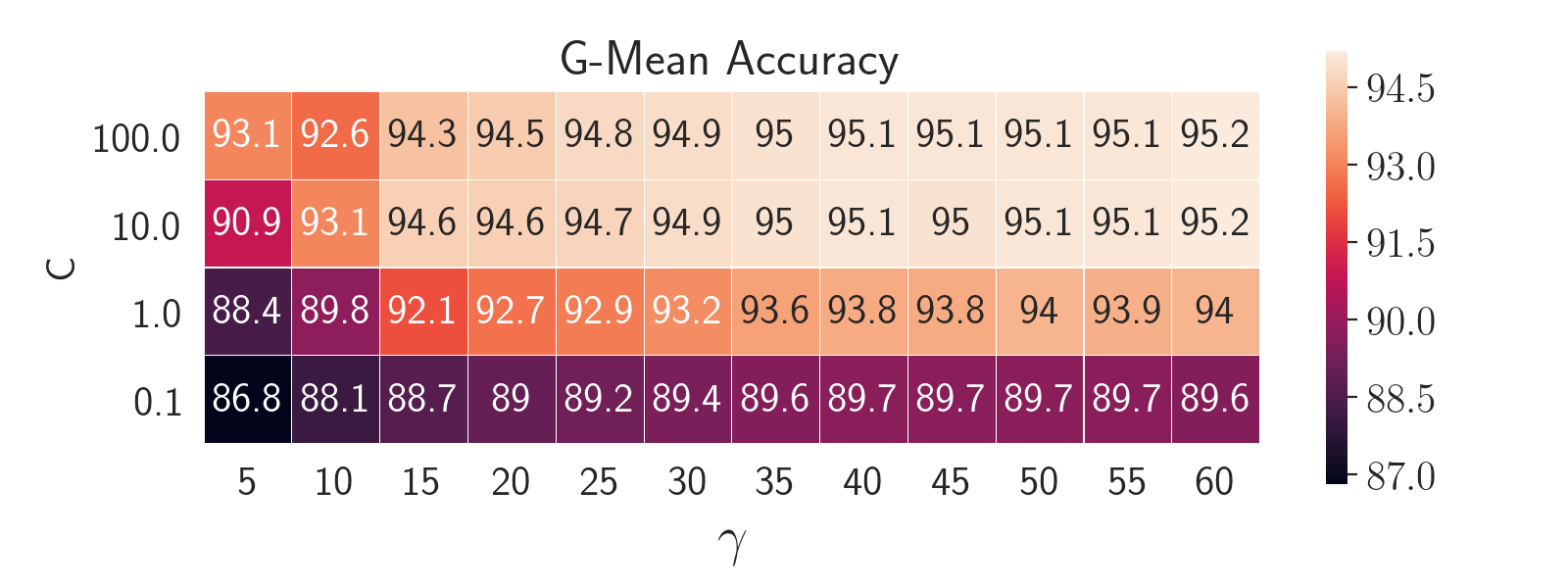}
        \caption{$T_e =8s$}
        \label{fig:svm8s}
    \end{subfigure}
    \caption{SVM parameter Tuning. In order to select the best set of parameters, we performed experiments for different combinations of the parameters $C$ and $\gamma$, and observed the G-Mean Accuracy.  These experiments were performed taking only a small portion of the data as training data and the rest as validation data, and they were repeated for $T_e$ = 2s, 5s, and 8s. Here it can be observed from three cases that as the episode length increased from 2s (\ref{fig:svm2s}) , to 5s (\ref{fig:svm5s}) , and even to 8s (\ref{fig:svm8s}) the performance improves. Thus the bigger windows are taken the better accuracy is obtained. For $T_e=5s$ and $T_e=8s$, the best results are obtained for $\gamma=45$ and $C=100$
    }
    \label{fig:svmParam}
\end{figure}

To tune the parameters we performed an exhaustive grid search. We tried out 3 different episode lengths, $T_e$ = 2 s, 5 s, and 8 s, as seen in literature. We trained the classifier using the training data and observed the performance on the validation data. The performance on the validation set is illustrated in Figure \ref{fig:svmParam}. Here we have shown the values of G-Mean Accuracy for different combinations of $\gamma$ and $C$, as it nicely balances both Sensitivity and Specificity. Please refer to the supplementary material where the other metrics have also been reported.

From the experiments we observe that as we increase the episode length, the performance improves.  G-Mean Accuracy improves from 88.244\% for $T_e=2$s to 91.966\% for $T_e=5$s and to 95.101\% for $T_e=8$s. Thus, an increase of $T_e$  by 3s roughly improves G-Mean accuracy by 4\%. From a practical consideration, we should set the episode length, $T_e$ such that the detection is both fast and accurate. However, there is a trade-off here: as we increase the episode length $T_e$ the detection accuracy improves but the computation becomes slower. So as a middle point we consider an episode length, $T_e = 5$s, for our further experiments.

Also, it can be observed that for $T_e=2s$, $C=1$ yields the best results, but for $T_e=5s$ and $T_e=8s$ the most promising results are obtained for $C \in [10,100]$. On the other hand, the best performing values of $\gamma$ are between 45 and 60. By performing some additional random experiments, it was seen that for $T_e=5s$, the set of parameter values $C=100$ and $\gamma=45$ consistently outperformed all other combinations. Thus, we selected them as the parameters of our SVM model.

\subsection{Feature Ranking by Random Forest Classifier}
\label{Feature Ranking by Random Forest Classifier}

\begin{figure}[h]
    \centering
    \begin{subfigure}[h]{0.49\textwidth}
        \includegraphics[width=\textwidth]{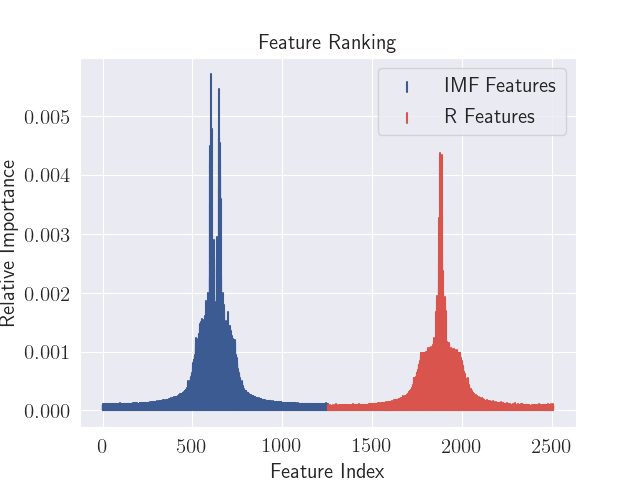}
        \caption{Relative Importance of the Features}
        \label{fig:ranking}
    \end{subfigure}    
    \hfill
    \begin{subfigure}[h]{0.49\textwidth}
        \includegraphics[width=\textwidth]{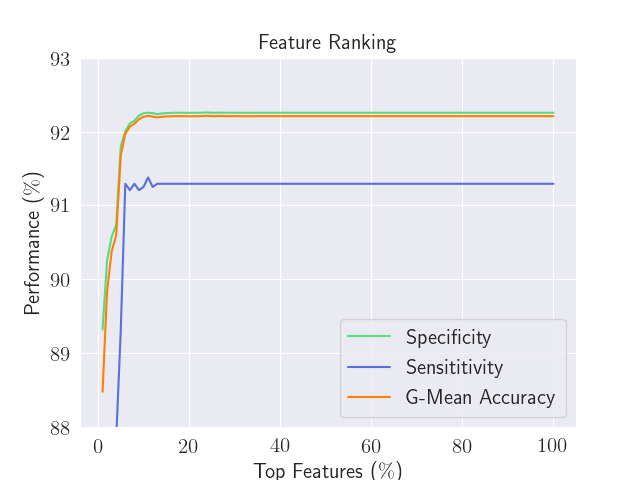}
        \caption{Effectiveness
        Different \% of Features}
        \label{fig:rankingRes}
    \end{subfigure}
    \caption{Feature Ranking using a Random Forest. In Figure \ref{fig:ranking} we have plotted the relative importance of the features. The features coming from IMF and R are colored in blue and red respectively. From the figure it is evident that a great deal of features is actually quite trifling, thus omission of them should make the classification more efficient. Thus, in Figure \ref{fig:rankingRes} we have shown the outcome of taking only a percentage of the top features, in terms of Specificity, Sensitivity, and G-Mean Accuracy. The performance gradually improves as we increase the number of features, but after taking $16 \%$ of the best features, it reaches a plateau. However, it was seen that taking the top $24 \%$ features yields a slight improvement of $0.002 \%$.}
    \label{fig:randFor}
\end{figure}

 We used a random forest comprising 750 decision trees to rank our features. After computing the feature importance using the random forest algorithm, we normalized the values and plotted them in Figure \ref{fig:ranking}. Here the first half consists of the IMF features (colored in blue) and the last half corresponds to the R features (colored in red).

Now, from the plot, we can observe that most of the features do not contribute at all in making an accurate prediction. These features originate from the DFT coefficients and since DFT coefficients for a real signal is symmetric, so are their significances. Moreover, the importance of the leftmost and the rightmost features of both IMF and R are negligible, they actually correspond to the high frequency components. The most impactful features are the features in the middle, and the importance gradually decreases as we move away from the center.


After ranking the features, we conduct our experiments on a reduced number of features based on their relative importance. We again take 3000 random samples from `VF' and 5000 random samples from `Not VF' as the training set and treat the rest as the validation set, as we did to tune the parameters. The ECG episodes are taken to be 5 seconds long, and the SVM parameters were selected as mentioned in the previous section.

We had selected the percentage of features to be considered, sorted by their importance and observed the Sensitivity, Specificity, and G-Mean accuracy on the validation set (Figure \ref{fig:rankingRes}). From the experiments, it was seen that only about top 16\% features were sufficient to sensibly distinguish between the two classes, and a slight improvement of 0.002\% was observed if we had taken the top 24\% features. Thus, we have selected the subset of only the top 24\% features, as our final feature set. 

This observation greatly reduces the dimensionality of the problem and makes both the training easier and the prediction faster. Also, this feature ranking exercise using Random Forest supports our hypothesis, that not all frequency components are needed for classification.

\subsection{Overcoming the Imbalance in the Dataset}
\label{Overcoming the Imbalance in the Dataset}
 As mentioned earlier, a major complication is that the dataset is highly imbalanced. This imbalance in the dataset prevents us from training our classifier properly. To overcome this limitation, we oversampled our data \cite{he2009learning}. We avoided random oversampling as it merely replicates some data from the minority class and as a result, often tends to overfit \cite{mease2007boosted}. Instead, we generated some synthetic data using Synthetic Minority Over-Sampling Technique (SMOTE) developed by Chawla et al. \cite{chawla2002smote} which avoids over-fitting by distributing the probability over the neighborhood of the minority class points in lieu of imposing too much bias on the given minority class points. Now with balanced SMOTE’d dataset, we train our machine learning classifier

\subsection{K-Fold Cross-Validation}
 Cross-Validation is an evaluation test that determines how well a model can generalize on an independent dataset. This removes any human biases that may have been introduced and also accounts for the variance in the dataset. Experiments on real-world datasets show that the best cross validation scheme is a 10-fold cross-validation \cite{kohavi1995study}. Thus we have performed a 10-fold cross validation on the balanced dataset, after random shuffling. We also performed a stratified 10-fold cross validation which maintains a uniform distribution of the classes \cite{kohavi1995study}. The results are summarized in table \ref{tab:res}.


\begin{table}[]
\begin{center}
\small
\caption{\label{tab:res}Experimental Results. Here we have shown both the results of the 10-Fold Cross Validation and Stratified 10-Fold Cross Validation tests.}
\makebox[\textwidth]{\begin{tabular}{|c|c|c|c|c|}
\hline
\multicolumn{5}{|c|}{10-Fold Cross Validation}                                                      \\ \hline
Data  & Sensitivity  & Specificity  & Accuracy  & G-Mean \\ & (\%) & (\%) & (\%) & Accuracy (\%) \\ \hline
Training Data & 100                & 100                & 100                & 100                  \\ \hline
Test Data     & 99.988 $\pm$ 0.016 & 98.401 $\pm$ 0.19  & 99.194 $\pm$ 0.092 & 99.191 $\pm$ 0.095   \\ \hline
\multicolumn{5}{|c|}{Stratified 10-Fold Cross Validation}                                           \\ \hline
Data  & Sensitivity  & Specificity  & Accuracy  & G-Mean \\ & (\%) & (\%) & (\%) & Accuracy (\%) \\ \hline
Training Data & 100                & 100                & 100                & 100                  \\ \hline
Test Data     & 99.992 $\pm$ 0.01  & 98.395 $\pm$ 0.187 & 99.194 $\pm$ 0.092 & 99.190 $\pm$ 0.096   \\ \hline
\end{tabular}}
\end{center}
\end{table}


Firstly, from the results, it can be observed that our proposed algorithm accurately classifies the training data. Thus, it may appear that the algorithm is overfitting the training dataset. However, it is not the case since our algorithm performs remarkably on the test data as well. The obtained value of Specificity is slightly higher in the ordinary 10-fold cross validation than in the stratified 10-fold cross validation. The opposite scenario is seen for Sensitivity. However, in both the cases both Sensitivity and Specificity is equally high, and it is reflected by the high value of G-Mean Accuracy. All these points to the effectiveness of our feature engineering and learning algorithm optimization.

It appears that our model predicts a number of false positives. But from further analysis it turned out that most of these false positives actually contained a small segment of VF within them, becoming VF after 1-2 seconds. Thus, these false positives are actually not harmful rather beneficial if we are to develop a real time predictive system.

\section{Comparison with Other Methods}
\label{Comparison}

Ventricular Fibrillation, being one of the most severe life threatening arrhythmias, is an exceedingly studied area. Research works have been undergoing in this area starting from the early `70s to even to date. In this section, we compare our work with some other well established works. During the study of other works, the following observations were apparent:
\begin{enumerate}
    \item Some researchers made a pre-selection of ECG signals by hand \cite{thakor1990ventricular,chen1987ventricular,kuo1978computer,barro1989algorithmic,zhang1999detecting,li1995detection,pan1985real,arafat2009detection,anas2011exploiting}. This resulted in better performance of their algorithms, but the accuracy drastically falls when tested on the entire dataset\cite{amann2005reliability}.
    \item All the datasets we have are imbalanced. It was observed in some research works that priority was given to the majority class (`Not VF') resulting in a higher Specificity and lower Sensitivity, albeit with a high accuracy \cite{anas2011exploiting,amann2005new,amann2007detecting,alonso2012feature,song2005support,asl2008support,acharya2017automated}.
    \item Surprisingly, a few authors tested their algorithms on the training set, resulting in a higher but clearly misleading accuracy \cite{song2005support}.
\end{enumerate}

Amann et al. \cite{amann2005reliability} presented a comparison of ten algorithms for the detection of VF. It was shown that none of the algorithms, namely, TCI \cite{thakor1990ventricular}, ACF \cite{chen1987ventricular}, VF Filter Method \cite{kuo1978computer}, Spectral Algorithm \cite{barro1989algorithmic}, Complexity Measure Algorithm \cite{zhang1999detecting}, Li Algorithm \cite{li1995detection}, Tompkins Algorithm \cite{pan1985real} etc., performed well over the entire dataset. The actual accuracy measures were very low compared to what was specified in the original papers since a pre-selection of the signals was done. Other algorithms in the literature that use only Signal Processing techniques also show a selectively high but overall poor performance. Arafat et al. \cite{arafat2009detection} presented a method based on EMD and Bayes Decision Theory, which shows a sensitivity (Se) of 99.00\% , specificity(Sp) of 99.88\% and accuracy(Ac) of 99.78\%, with only `VF' and `NSR' (Normal Sinus Rhythm) signals in the dataset. But the performance falls drastically when other beats and rhythms are considered. Anas et al. \cite{anas2011exploiting} presented another algorithm based on the EMD which obtains a sensitivity of 82.89\%, specificity of 99.02\% and accuracy of 98.62\%, on the test set. The authors not only made a pre-selection of the signals but also gave priority to the Not VF class while selecting the threshold which results in a high accuracy but low sensitivity. Two other algorithms, Hilbert transform (HILB) \cite{amann2005new} and Phase Space Reconstruction (PSR) \cite{amann2007detecting} algorithms use the phase space of the ECG signal for VF detection. But they don’t consider the shape of this signal. Thus, they fail to differentiate VT (Ventricular Tachycardia) from VF when other arrhythmias are present (HILB: Se = 79.73, Sp =98.83, Ac = 98.40; PSR: Se = 78.07, Sp= 99.01, Ac = 98.53).

Machine learning approaches show great promises as they improve the detection accuracy. Almost all the machine learning algorithms in the literature extract features from other existing VF detection algorithms based on signal processing, and then use them as features to detect VF. Atienza et al. \cite{alonso2012feature} studied a number of ECG signal parameters and used them as features to train an SVM model (Se=74.1, Sp=94.7). In a later work \cite{alonso2014detection}, they considered ECG segments of 8s long to compute temporal and spectral parameters as features and then they developed a classifier using SVM. This obtained an accuracy of 91.1\% while detecting VF. Qiao et al. \cite{li2014ventricular} presented quite a similar pipeline; some parameters were computed as features and a genetic algorithm was used to rank the features. Then, weights were set to classes to consider the imbalance in the dataset and finally, an SVM model was trained (Se=96.2\%, Sp=96.2\%, Ac=96.3\%). However, a drawback of the algorithm is that it considered Ventricular Fibrillation, Ventricular Tachycardia, and Ventricular Flutter all as the VF class. As these three classes are quite similar in appearance, an algorithm should be robust enough to distinguish among these three. Verma et al. \cite{verma2016detection} used similar approaches to extract features and subsequently used a Random Forest classifier to detect VF signals. This algorithm obtained Ac = 94.79, Se = 95.04, Sp = 94.78 for 5s long episodes and better results for 8s long episodes (Ac = 97.20, Se = 95.05, Sp = 97.02). Mi et al. \cite{song2005support} experimented with different dimensionality reduction and machine learning algorithms. 
 Strangely, their models were tested on the training set, still, it obtained poor sensitivity (Se = 92.396, Sp = 99.121, Ac = 99.350). Asl et al. \cite{asl2008support} did some preprocessing on the ECG signals before extracting the features and performing a dimensionality reduction using Gaussian Discriminant Analysis (GDA); finally, an SVM model was trained. This algorithm is by far the best performing one (Se = 95.77, Sp = 99.40, Ac = 99.16), but the drawback is that it works on the window length of 32 R-R interval, which is roughly 30 seconds. This makes the algorithm extremely slow and delays the prediction. Clayton et al. \cite{clayton1994recognition} used a (shallow) neural network to classify `VF', which unfortunately obtained poor performance (Se = 86, Sp = 58). Acharya et al. \cite{acharya2017automated} very recently used a Convolutional Neural Network (CNN) to detect VF along with some other arrhythmias. Their algorithm obtained a high specificity (98.19) but very low sensitivity (56.44).

In Table \ref{tab:comp} we present a comparison among a number of methods. Here we are only showing the algorithms based on machine learning techniques as they outperform traditional signal processing based algorithms. Note that the results reported in Table \ref{tab:comp} should be interpreted carefully as different works have reported their results using different validation techniques. As has already been mentioned, in our work, we employed 10-fold cross validation, and avoided independent testing as the idea of the latter has been doubted by some researchers (e.g., \cite{chou2011some}). More specifially, as argued by Chou in \cite{chou2011some}, the way of independent test instance selection to test the predictor could be quite arbitrary resulting in arbitrary conclusions. A predictor achieving a higher success rate than the other predictor for a given independent testing dataset might fail to keep so when tested by another independent testing dataset. Accordingly, independent testing is not a fairly objective test method although it has often been used to demonstrate the practical application of a predictor in different domains. Unfortunately most of the works we are comparing with, employed  independent testing but without any specific information about the test dataset; also they have not published the code of their implementation making it impossible to reproduce the test results and ensure a level playing field. 
	
	\begin{table}[h]
		\caption{Comparison among different methods. Results are taken from the respective papers. `-' indicates that the score was not available/reported in the respective paper. Here, we have only included the machine learning based works, as the outperforms the traditional approaches based solely on signal processing (as shown in section \ref{Comparison}.)}
		\label{tab:comp}
		\small
        \begin{tabular}{ | c | c | c | c | c |}
			\hline
			Algorithm & Sensitivity & Specificity & Accuracy & G-Mean \\  & (\%) & (\%) & (\%) & Accuracy (\%)\\
			\hline
			Atienza et al. \cite{alonso2012feature} & 74.1 & 94.7 & - & 83.77 \\
			\hline
			Atienza et al. \cite{alonso2014detection} & - & - & 91.1 & - \\
			\hline
			Qiao et al. \cite{li2014ventricular} & 96.2 & 96.2 & 96.3 & 96.2 \\
			\hline
			Verma et al. \cite{verma2016detection} & 95.04 (5s) & 94.78 (5s) & 94.79 (5s) & 94.91 (5s)  \\
             & 95.05 (8s) & 97.02 (8s) & 97.20 (8s) & 96.03 (8s) \\
			\hline
			Mi et al. \cite{song2005support} & 92.396 & 99.121 & 99.350 & 95.70 \\
			\hline
			Asl et al. \cite{asl2008support} & 95.77 & 99.40 & 99.16 & 97.57 \\
			\hline
			Clayton et al. \cite{clayton1994recognition} & 86 & 58 & - & 70.63 \\
			\hline
			Acharya et al. \cite{acharya2017automated} & 56.44 & 98.19 & 97.88 & 74.44 \\
			\hline \hline
			VFPred & $99.988 \pm 0.016$ & $98.401 \pm 0.19$ & $99.194 \pm 0.092$ & $99.191 \pm 0.95$ \\
			\hline
			
		\end{tabular}

	\end{table}

\section{Conclusion}
Ventricular Fibrillation is a dangerous life threatening arrhythmia that can cause sudden cardiac arrest resulting in a sudden death. The inability of a doctor to continuously monitor the heart conditions of all the patients motivates us to design and develop efficient and accurate automated systems to perform the task. Immediately after Ventricular Fibrillation is detected a shock treatment can be given to the patient. A lot of research works have been done to detect Ventricular Fibrillation using either signal processing or machine learning techniques separately. In some works, the authors even proposed automated systems to give shock treatment to the patients. They argued that the system should have higher specificity, i.e., the accuracy of detecting patients not affected by Ventricular Fibrillation should be higher. This imposes priority on detecting `Not VF' class correctly and in most cases the accuracy of detecting `VF' was low. We find this contradictory, as the objective is to detect VF, not the other. Moreover, such a system would be vulnerable if it fails to detect VF consistently which may result in deaths of patients. We do not deny the importance of high specificity but at the same time, we believe that the sensitivity should be high as well. Note that, our goal is to aid the doctors and in no way to replace them. Hence, instead of developing automated shocking systems that fail in detecting ‘VF’ properly with the excuse of preventing the shock treatment of unaffected patients, we propose an automated monitoring system which will continuously monitor the patients and when the possibility of `VF’ arises, it will activate an alarm and an experienced doctor would examine the case and decide whether shock treatment is indeed required. Thus the sensitivity of the detection/prediction system should be high to avoid unfortunate deaths. We also gave importance to specificity, as our algorithm can classify both the classes with near equal performance.

In this work, we combined the strengths of both signal processing and machine learning. We analyzed the pattern of the `VF' class ECG signals and proposed a novel feature engineering scheme. Next, we trained machine learning classifiers on the extracted features. As the dataset is highly imbalanced we adopted appropriate techniques to overcome the imbalance. Also, we computed feature importance and experimentally found that similar performance can be obtained by using a much smaller subset of the computed features, which can be very helpful for constructing a real time system as it would make our algorithm much faster. We have compared our algorithm with the state of the art and found out that our algorithm outperforms them. The only algorithm that comes close to our algorithm is presented by Asl et al \cite{asl2008support}. However, the downside of their algorithm is that it works on windows of 32 R-R intervals ( roughly 30 seconds), obtaining a sensitivity of 95.77\%, specificity of 99.40\%, accuracy of 99.16\% and G-Mean Accuracy of 97.57\%. On the other hand, our algorithm obtains a sensitivity of 99.99\%, specificity of 98.40\%, accuracy of 99.20\% and G-Mean Accuracy of 99.19\%, on a small 5 seconds long window. It was seen experimentally that our performance improves as we increase the window length.

In conclusion, this work has shown that the combination of signal processing based feature engineering and machine learning based decision making can greatly improve the performance of algorithms in biomedical signal processing domain. While existing algorithms emphasize either on signal processing or on machine learning, leaving the other one neglected, our objective was to make a fusion of these two approaches to take the best of both techniques. Thus we have developed a robust algorithm, VFPred, for the detection of Ventricular Fibrillation that is both fast and accurate. One of the most noteworthy features of VFPred is that it can classify both the classes equally accurately, even when using a short ECG signal. This is a significant improvement as existing works, while can identify the majority class (`Not VF') very well, fall much short in identifying the minority class (`VF'). Also, the success of VFPred with short ECG signals opens up the possibilities for developing more responsive, real time cardiac monitoring systems.

\section{Conflict of Interests}
The authors declare that there is no conflict of interests.

\section{Acknowledgments}
This research was funded by ICT Division-Government of the People's Republic of Bangladesh.





\bibliographystyle{unsrt}
\bibliography{bibfile}

\end{document}